\documentclass{article}
\usepackage{spconf,amsmath,graphicx,hyperref}
\usepackage{xcolor, colortbl}
\usepackage[table]{xcolor}
\usepackage[dvipsnames]{xcolor}
\usepackage{booktabs}
\usepackage{bbding}
\usepackage{multirow}

\usepackage{xurl}

\usepackage[most]{tcolorbox}
\usepackage{colortbl}
\tcbuselibrary{breakable} %breakable：支持跨页

\title{MAC-SLU: Multi-Intent Automotive Cabin Spoken Language Understanding Benchmark}

\name{
    \begin{tabular}{c} % 使用表格环境包裹，c 表示居中对齐
        Yuezhang Peng\textsuperscript{1}, Chonghao Cai\textsuperscript{1}, Ziang Liu\textsuperscript{2}, Shuai Fan\textsuperscript{3}, Sheng Jiang\textsuperscript{3}, Hua Xu\textsuperscript{3}, Yuxin Liu\textsuperscript{1}, \\ 
        \textit{Qiguang Chen\textsuperscript{2}, Kele Xu\textsuperscript{4}, Yao Li\textsuperscript{5}, Sheng Wang\textsuperscript{1,5}, Libo Qin\textsuperscript{2,*}, Xie Chen\textsuperscript{1,*}\thanks{* Corresponding authors.}}
    \end{tabular}
}
    
\address{
    \textsuperscript{1} School of Computer Science, MoE Key Lab of Artificial Intelligence, Shanghai Jiao Tong  University
    \\ \textsuperscript{2}School of Computer Science and Engineering, Central South University \\ 
    \textsuperscript{3}AISpeech Co., Ltd. 
    \textsuperscript{4} National University of Defense Technology, \textsuperscript{5}Shanghai Aviation Electric Co., Ltd
  }

\begin{document}
 
\ninept
% \fontsize{9.5pt}{11pt}\selectfont
%
\maketitle
\begin{abstract}

Spoken Language Understanding (SLU), which aims to extract user semantics to execute downstream tasks, is a crucial component of task-oriented dialog systems. Existing SLU datasets generally lack sufficient diversity and complexity, and there is an absence of a unified benchmark for the latest Large Language Models (LLMs) and Large Audio Language Models (LALMs). This work introduces MAC-SLU, a novel Multi-Intent Automotive Cabin Spoken Language Understanding Dataset, which increases the difficulty of the SLU task by incorporating authentic and complex multi-intent data. Based on MAC-SLU, we conducted a comprehensive benchmark of leading open-source LLMs and LALMs, covering methods like in-context learning, supervised fine-tuning (SFT), and end-to-end (E2E) and pipeline paradigms. Our experiments show that while LLMs and LALMs have the potential to complete SLU tasks through in-context learning, their performance still lags significantly behind SFT. Meanwhile, E2E LALMs demonstrate performance comparable to pipeline approaches and effectively avoid error propagation from speech recognition. Code\footnote{\url{https://github.com/Gatsby-web/MAC\_SLU}} and datasets\footnote{\url{huggingface.co/datasets/Gatsby1984/MAC\_SLU}} are released publicly.

\end{abstract}
\section{Introduction}
\label{sec:intro}

Spoken Language Understanding (SLU) is a conventional paradigm for task-oriented spoken semantics extraction, which has been widely applied in scenarios such as smart homes and automobiles to extract spoken commands from users for executing downstream tasks~\cite{tur2011spoken}. Traditional SLU employs a pipeline approach, first transcribing the user's spoken query into text via Automatic Speech Recognition (ASR), and then extracting semantic information through Natural Language Understanding (NLU), which typically includes Intent Classification (IC) and Slot Filling (SF)~\cite{mesnil2014using}. Subsequently, E2E SLU systems emerged to address the issue of error propagation from the ASR transcription process~\cite{lugosch2019speech} and potentially incorporate pre-trained language models (e.g., RoBERTa~\cite{liu2019roberta}) to enhance performance~\cite{cheng2023ml}. Presently, the advancement of LLMs and LALMs offers the potential for a more flexible and precise SLU.

Nevertheless, research on LLM-based SLU still confronts the following two challenges: (1) Existing SLU datasets lack sufficient diversity and complexity. The widely used ATIS~\cite{hemphill1990atis} and SNIPS~\cite{coucke2018snips} datasets contain only 16 and 7 intents, respectively, which limits the task's difficulty, resulting in existing models already achieving over 95\% accuracy in both IC and SF~\cite{qin2021co}. SLURP dataset~\cite{bastianelli2020slurp} increases the number of intent and slot categories but remains confined to single-intent SLU tasks. (2) There is a lack of a unified benchmark for state-of-the-art (SOTA) open-source LLMs and LALMs. Although some studies have preliminarily explored the performance of LLMs like ChatGPT and Llama on certain SLU tasks~\cite{he2023can,li2024whisma}, they have adopted different task formats (i.e., varying data formats, prompts, or training and alignment methods), leading to evaluation results that cannot be fairly compared. Furthermore, existing research has been limited to pipeline-based methods and LLMs, without exploring E2E approaches and the most advanced LALMs.

% In this paper, we release the \textbf{Multi-intent Automotive Cabin Spoken Language Understanding Dataset} (MAC-SLU), a novel Chinese multi-intent SLU dataset, and conduct an extensive benchmark of open-source LLMs and LALMs to address the two aforementioned challenges. The MAC-SLU dataset is derived from real-world automotive cabin text command data, with speech synthesized via a SOTA Text-to-Speech (TTS) model. It encompasses 8 domains and 81 intents, and includes multi-intent data with up to 5 intents, surpassing existing SLU datasets in terms of diversity and complexity. Based on this dataset, we establish a unified benchmark for open-source LLMs and LALMs. Specifically, our contributions include:

This work addresses these challenges with two steps. First, we introduce the \textbf{Multi-Intent Automotive Cabin Spoken Language Understanding} (\textbf{MAC-SLU}) dataset, a novel Chinese SLU corpus to overcome the complexity limitations of existing data. Derived from real-world automotive text commands with TTS-synthesized speech, it spans 8 domains, 81 intents, 192 slots, and includes multi-intent queries with up to 5 intents, creating a more rigorous testbed. Second, we establish a unified benchmark on MAC-SLU for SOTA open-source LLMs and LALMs. Standardizing formats, tasks, and evaluation methods enables fair comparisons and provides a dependable reference for the community. Our contributions include:

\begin{itemize}
    % \item We release a novel Chinese multi-intent SLU dataset, creating a more challenging task for LLM-based SLU research.
    \item We introduce MAC-SLU, a novel Chinese multi-intent SLU dataset including complex, multi-intent queries from a real-world automotive cabin domain. MAC-SLU enables a more chanllenging evaluation of the latest LLMs and LALMs.

    \item We provide a comprehensive benchmark for SOTA open-source and closed-source LLMs and LALMs, encompassing methods based on direct inference, in-context learning, and SFT, as well as both pipeline and E2E SLU task paradigms.

    \item Our experiments demonstrate that (1) existing LLMs and LALMs can complete parts of the IC or SF tasks through in-context learning, yet there remains a significant performance gap compared to in-domain SFT; (2) benefiting from the avoidance of error propagation, current LALMs can already achieve performance comparable to pipeline methods.
\end{itemize}

\section{Related Work}
\label{sec:related}

% \begin{table*}[htbp]
%     \centering
%     \begin{tabular}{lccccccc} 
%         \toprule % 表格顶部粗线
%          Dataset & \textbf{ATIS} & \textbf{SNIPS} & \textbf{FSC} & \textbf{SLURP} & \textbf{MixATIS} & \textbf{MixSNIPS} & \textbf{MAC-SLU} \\ \midrule
%         Domain & 1 & 2 & 2 & \textbf{18} & 1 & 2 & 8 \\
%         Intent & 16 & 7 & 6 & 46 & 16 & 7 & \textbf{81} \\
%         Slot & 41 & 4 & 2 & 56 & 41 & 4 & \textbf{192} \\
%         Multi-intent & \XSolidBrush & \XSolidBrush & \XSolidBrush & \XSolidBrush & \Checkmark & \Checkmark & \Checkmark \\
%         \bottomrule % 表格底部粗线
%     \end{tabular}
%     \caption{Comparison of SLU datasets. MAC-SLU supports the largest intent/slot categories and complex multi-intent data.}
%     \label{tab:dataset_comparison}
% \end{table*}

\subsection{SLU Datasets}

\begin{table*}[ht]
    \centering
    \renewcommand{\arraystretch}{1.1} % Sets the row spacing
    
    \begin{tabular}{l|ll}
        \toprule
        % -- Header Row --
        \textbf{Domain} & \textbf{Intent} & \textbf{Slot} \\
        \hline
        
        % -- Data Rows --
        Car Control & Car System Control, Car Body Control & Object, Location, Function, Application \\
        Map & Navigation, Provide Address, Check Traffic & Destination Name, Destination Type, Waypoint Name \\
        Music & Play Music, Query Music Information & Music Name, Music Type, Artist Name \\
        \bottomrule
    \end{tabular}
    \caption{Examples of domains, intents, and slots in the MAC-SLU dataset.  All data shown are English translations of the original Chinese.}
    \label{tab:domain_intent_entity}
\end{table*}

Existing SLU datasets primarily consist of single-intent datasets such as ATIS, SNIPS, FSC, and SLURP \cite{hemphill1990atis,coucke2018snips,lugosch2019speech, bastianelli2020slurp}, as well as multi-intent datasets like MixATIS and MixSNIPS \cite{qin2020agif}. The ATIS dataset, released in 1990, mainly contains voice queries for flight information and includes only 16 intent categories. More recent datasets like SNIPS and FSC are designed for smart home scenarios but also have a limited number of intent classes. Another commonly used large-scale SLU dataset is SLURP, which expands the number of intent categories to 46, significantly increasing the task's complexity. Nevertheless, SLURP does not contain multi-intent data, which limits its diversity. The MixATIS and MixSNIPS multi-intent datasets were constructed from the ATIS and SNIPS datasets, respectively, by connecting sentences with different intents using conjunctions (e.g.,``and"). However, constrained by the relative simplicity of the original ATIS and SNIPS datasets, the intent diversity in MixATIS and MixSNIPS is also limited. Furthermore, E2E Chinese SLU datasets are scarce, and the closest alternatives, CAIS \cite{liu2019cm} and ECDT-NLU\footnote{\url{conference.cipsc.org.cn/smp2019/evaluation.html}}, are text-based datasets for NLU tasks.

\begin{table}[t]
    \centering
    % 在列定义 l 后面加上 | 来添加竖线
    % 在表格外侧也加上 | 来封闭表格
    % 使用 \hline 代替 booktabs 的线来确保连接
    \begin{tabular}{l|cccc}
        \toprule
         \textbf{Dataset} & \textbf{Domain} & \textbf{Intent} & \textbf{Slot} & \textbf{Multi-intent} \\ \hline
        \textbf{ATIS} \cite{hemphill1990atis} & 1 & 16 & 41 & \XSolidBrush \\
        \textbf{SNIPS} \cite{coucke2018snips} & 2 & 7 & 4 & \XSolidBrush \\
        \textbf{FSC} \cite{lugosch2019speech} & 2 & 6 & 2 & \XSolidBrush \\
        \textbf{SLURP} \cite{bastianelli2020slurp} & \textbf{18} & 46 & 56 & \XSolidBrush \\
        \textbf{MixATIS} \cite{qin2020agif} & 1 & 16 & 41 & \Checkmark \\
        \textbf{MixSNIPS} \cite{qin2020agif} & 2 & 7 & 4 & \Checkmark \\ \hline
        \textbf{MAC-SLU} & 8 & \textbf{81} & \textbf{192} & \Checkmark \\
        \bottomrule
    \end{tabular}
    \caption{Comparison of SLU datasets. MAC-SLU supports the largest intent/slot categories and complex multi-intent data.}
    \label{tab:dataset_comparison}
\end{table}

% Consequently, existing datasets are still confined to a small number of intent categories and single-intent data. There is a lack of an E2E SLU dataset that covers a large number of domains and intent categories and includes realistic multi-intent data. 

\subsection{LLMs for SLU}

% LLMs trained based on scaling laws have demonstrated emergent reasoning and generalization capabilities, enabling them to solve a wide range of question-answering, mathematical, or coding problems \cite{ouyang2022training}. 

Several studies have attempted to apply LLMs to SLU tasks. He \cite{he2023can} explored using in-context learning to enable ChatGPT to perform IC and SF, demonstrating that LLMs can achieve commendable performance in IC but still face challenges with the more complexly defined SF task. Yin \cite{yin2024large} utilized an LLM fine-tuned with LoRA \cite{hu2022lora} for SLU tasks and achieved superior performance compared to methods trained from scratch or fine-tuned on masked language models. WHISMA \cite{li2024whisma}, which employs a Whisper encoder \cite{radford2023robust} and a Llama-3 decoder \cite{dubey2024llama} and is trained with modal alignment, is capable of performing zero-shot SLU tasks and has surpassed pipeline-based methods. Although these approaches all leverage LLMs for SLU tasks, they use different models, training methods, and data, making it impossible to compare them within a unified benchmark.

\section{MAC-SLU Dataset}\label{sec:mac-slu}

In this section, we provide details on text data collection and speech data synthesis, and a specific dataset analysis for MAC-SLU.

\subsection{Text Data Collection}

\begin{table}[t]
    \centering
    \renewcommand{\arraystretch}{1.2} % 设置表格行间距
    
    \begin{tabular}{c|cccc|c}
        \toprule
        % -- 表头行 --
        \textbf{Intents} & \textbf{Train} & \textbf{Dev} & \textbf{Test} & \textbf{Overall} & \textbf{Ratio (\%)} \\
        \hline
        
        % -- 表格数据行 --
        0 & 5305 & 419 & 26 & 5750 & 28.00 \\
        1 & 10018 & 768 & 826 & 11612 & 56.54 \\
        2 & 2183 & 166 & 246 & 2595 & 12.63 \\
        3 & 379 & 26 & 50 & 455 & 2.22 \\
        $\geq$4 & 112 & 12 & 3 & 127 & 0.62 \\
        \hline
        Total & 17997 & 1391 & 1151 & 20539 & 100.00 \\
        
        \bottomrule
    \end{tabular}
    
    % 您可以根据需要修改表格的标题和标签
    \caption{Distribution of samples by the number of intents in the train, dev, and test sets of MAC-SLU dataset, with totals and ratios.}
    \label{tab:my_separated_ratio_table}
\end{table}

The text data for the MAC-SLU dataset originates from real-world automotive cabin scenarios. We collected over 20,000 transcribed texts of Chinese spoken commands, along with their corresponding SLU parsing results. The training and validation sets consist of 17,997 and 1,391 samples, respectively, which were randomly selected and directly utilized the existing SLU parsing results as weakly labeled data. For the test set, we randomly partitioned 1,800 samples from the dataset. These samples were then manually reviewed and curated by three data annotators, who removed or corrected a small number of samples with parsing errors or blank intents. This process resulted in a final clean test set of 1,152 samples.

\subsection{Speech Data Generation}

We synthesized the corresponding Mandarin Chinese speech data for the transcribed texts using CosyVoice-2 \cite{du2024cosyvoice}. The speaker embedding templates for the TTS process were derived from AIShell-1 \cite{bu2017aishell}, a widely used Chinese ASR dataset. Specifically, for each speaker in the train, dev, and test sets of AIShell-1, we randomly selected audio clips to create three distinct sets of speaker templates. When synthesizing each audio sample for MAC-SLU, a template was randomly chosen from the corresponding set to ensure speaker diversity and maintain isolation across different data splits.

\subsection{Dataset Analysis}

We compared the semantic diversity of MAC-SLU with commonly used datasets, including ATIS, SNIPS, FSC, and SLURP in Table \ref{tab:dataset_comparison}. MAC-SLU is surpassed by SLURP only in the richness of its domains. This is primarily because MAC-SLU focuses on domains within automotive cabins, where the range of tools a user can invoke is relatively limited. In contrast, the smart-home scenario targeted by SLURP often involves a greater number of callable tools across more domains. However, in terms of the variety of intents and slots, MAC-SLU exceeds all existing single and multi-intent datasets, supporting more fine-grained intent and entity classification.

\begin{table*}[htbp]
    \centering
    % -- 用户提供的设置 --
    \setlength{\cmidrulewidth}{0.05mm} % 设置cmidrule粗细
    \renewcommand{\arraystretch}{1.2} % 适当增加了行间距以适应新结构
    % --------------------
    
    \begin{tabular}{ll p{2mm} ccc p{2mm} ccc p{2mm} ccc}
        \toprule
        % -- 主标题行 --
        & & & \multicolumn{3}{c}{\textbf{0-shot}} && \multicolumn{3}{c}{\textbf{5-shot}} && \multicolumn{3}{c}{\textbf{10-shot}} \\
        \cmidrule(lr){4-6} \cmidrule(lr){8-10} \cmidrule(lr){12-14}
        
        % -- 副标题行 --
        \textbf{Category} & \textbf{Model} && \textbf{Acc} & \textbf{F1} & \textbf{OA} && \textbf{Acc} & \textbf{F1} & \textbf{OA} && \textbf{Acc} & \textbf{F1} & \textbf{OA} \\
        \midrule
        
        % -- NLU 部分 --
        \multirow{4}{*}{\textbf{NLU}} 
        & Qwen3-1.7B && 25.10 & 13.97 & 0.17 && 47.61 & 30.85 & 3.13 && 58.64 & 34.57 & 1.91 \\
        & Qwen3-4B   && 30.06 & 0.20 & 0.87 && 63.16 & 43.32 & 6.08 && 60.64 & 45.44 & 7.65 \\
        & Qwen3-8B   && 38.75 & 1.74 & 1.22 && 59.86 & 46.82 & 9.38 && 65.42 & 49.50 & 10.69 \\
        & Qwen3-32B  && 67.07 & 0.70 & 0.52 && 70.11 & 51.76 & 14.16 && \textbf{70.37} & \textbf{55.09} & \textbf{14.42} \\
        \midrule

        % -- ASR + NLU 部分 --
        % \multirow{4}{*}{\textbf{Pipeline (ASR + NLU)}} 
        \multirow{4}{*}{\textbf{\begin{tabular}[l]{@{}l@{}}Pipeline \\ (ASR + NLU)\end{tabular}}} 
        & Whisper + Qwen3-1.7B && 24.07 & 3.24 & 0.09 && 46.83 & 27.22 & 2.09 && 56.30 & 31.00 & 1.65 \\
        & Whisper + Qwen3-4B   && 26.24 & 0.13 & 0.87 && 58.47 & 36.86 & 5.56 && 55.26 & 39.17 & 6.86 \\
        & Whisper + Qwen3-8B   && 35.88 & 1.47 & 1.48 && 54.91 & 39.84 & 8.43 && 60.73 & 42.72 & 9.30 \\
        & Whisper + Qwen3-32B  && 61.69 & 0.70 & 0.47 && 65.51 & 43.62 & 9.73 && \textbf{66.12} & \textbf{47.38} & \textbf{10.86} \\
        \midrule
        
        % -- E2E SLU 部分 --
        \multirow{5}{*}{\textbf{E2E SLU}} 
        & Qwen2-Audio-Instruct && - & - & - && - & - & - && - & - & - \\
        & Qwen2.5-Omni-3B && 27.28 & 2.73 & 2.09 && 40.31 & 24.48 & 4.08 && 34.58 & 27.84 & 4.52 \\
        & Qwen2.5-Omni-7B && 30.67 & 6.20 & 1.39 && 58.73 & 39.63 & 7.73 && \textbf{62.47} & 43.79 & 8.95 \\
        & Phi-4-Multimodal-Instruct && 16.5 & 3.44 & 0.70 && 17.98 & 16.15 & 2.95 && 22.76 & 21.40 & 2.87 \\
        & GPT-4o-Audio && 50.12 & 1.14 & 2.31 && 52.53 & 44.32 & 11.34 && 55.92 & \textbf{46.45} & \textbf{12.21} \\
        & Gemini-2.5-Flash && 48.91 & 18.20 & 2.52 && 42.31 & 36.36 & 10.43 && 45.61 & 38.34 & 10.86 \\
        \bottomrule
    \end{tabular}
    \caption{In-context learning results for LLMs and LALMs on MAC-SLU dataset. Acc, F1, and OA represent Intent Classification Accuracy, Slot Filling F1-Score, and Overall Accuracy, respectively. The best results in each category are shown in \textbf{bold}. Qwen2-Audio-Instruct exhibited weak instruction-following abilities and failed to produce outputs in our required format.}
    \label{tab:in_context_learning}
\end{table*}

We present examples of domains, intents, and slots from the MAC-SLU dataset in Table \ref{tab:domain_intent_entity} to facilitate an understanding of the specific data categorization. Due to space constraints, only a selection of examples is provided. The remaining domains include phone call, radio, weather, movies, and playback control.

\section{Experiments}
\label{sec:exp}

In this section, we evaluate the performance of existing LLMs and LALMs on the MAC-SLU dataset with direct inference, in-context learning, and SFT methods.

\textbf{Models. }Our experiments primarily target open-source LLMs and LALMs. For the LLMs experiments, we used various sizes of the Qwen3 \cite{yang2025qwen3}. For the multi-modal experiments, we investigated Qwen2-Audio-Instruct \cite{chu2024qwen2}, Qwen2.5-Omni \cite{xu2025qwen2}, Phi-4-multimodal-instruct \cite{abouelenin2025phi}, and MiniCPM-o-2\_6 \cite{yao2024minicpm}. For the ASR model, we employed Whisper-Large-V3-Turbo \cite{radford2023robust} or Paraformer \cite{gao2022paraformer}, which achieves a Character Error Rate (CER) of 10.40\% or 3.64\% on the MAC-SLU test set, respectively. As a supplement, we also tested the more powerful closed-source model, GPT4o-Audio-Preview-2024-12-17 and Gemini-2.5-Flash.

\textbf{Implementation Details. }Our experiments were conducted in two parts. For in-context learning methods, all experiments were run on Nvidia H20 GPUs, with inference accelerated using vLLM \cite{kwon2023efficient} deployment. For SFT methods, all experiments were conducted on Nvidia 3090 GPUs. We uniformly adopted the Llama-Factory \cite{zheng2024llamafactory} training framework and applied the LoRA \cite{hu2022lora} method to implement parameter-efficient fine-tuning. The LoRA rank and alpha were set to 16 and 32, respectively, following common default settings. 

% The fine-tuning objective for SFT was to generate complete intent and slot parsing results, so the results support the calculation of IC accuracy, SF F1 score, and Overall accuracy.

% As Llama-Factory does not currently support the fine-tuning of Phi-4-multimodal-instruct, we fine-tuned the Phi model separately by adapting the code from \textcolor{red}{[citation]}. The batch size per step was set to 2, and we used 8 steps of gradient accumulation to simulate a batch size of 16.

\textbf{Evaluation Metrics. }We followed the standard metrics for SLU tasks \cite{qin2021survey}. For the IC task, we calculated accuracy, and for the SF task, we calculated the F1 score. The Overall accuracy is calculated as the probability that IC and SF tasks are simultaneously correct.

% \subsection{In-Context Learning Results}

% \begin{figure}[ht!]
%     \centering
%     \begin{tcolorbox}[
%         colback=black!5!white,
%         colframe=black,
%         boxrule=0.8pt,
%         left=0.5mm,
%         right=0.5mm,
%         top=0.5mm,
%         bottom=0.5mm
%     ]
%     \begin{lstlisting}[basicstyle=\tiny\ttfamily, breaklines=true]
% DOMAIN_INTENT_LIST =
% - Car Control: Car System Control, Car Body Control
% - Map: Navigation, Provide Address, Check Traffic
% [...]

% SLOT_LIST = 
% - Car Control-Object: Specific objects for car control, for example, air conditioning, interior lights, reading lights.
% - Map-Map Size: The zoom level of the map, for example, minimum, zoom out.
% [...]

% SYSTEM_PROMPT_TEMPLATE =
% You are an expert in NLU for automotive cabin systems.
% Your task is to identify all the slots and their corresponding values from the user's query for slot filling.

% You need to follow these rules:
% 1.  Identify Multiple Semantic Frames: A single user query may contain multiple independent intents. You need to generate a corresponding semantic structure for each intent.
% 2.  Strict Matching: The identified domain, intent, and slot names must be strictly selected from the list below. 

% [...]

% Available Domain-Intent List:
% {DOMAIN_INTENT_LIST}

% Available Slot List:
% {SLOT_LIST}
%     \end{lstlisting}
%     \end{tcolorbox}
%     \caption{In-context learning prompt template for jointly SLU task. The intent lists, slot lists and format are partially omitted for brevity.}
%     \label{fig:my_prompt_example}
% \end{figure}

\begin{figure}[ht!]
    \centering
    \begin{tcolorbox}[
        colback=black!5!white,
        colframe=black,
        boxrule=0.8pt,
        left=0.5mm,
        right=0.5mm,
        top=0.5mm,
        bottom=0.5mm
    ]
    % -- 主要修改在这里 --
    \begin{lstlisting}[
        basicstyle=\tiny\ttfamily, 
        breaklines=true,
        morekeywords={DOMAIN_INTENT_LIST, SLOT_LIST, SYSTEM_PROMPT_TEMPLATE}, % 添加需要高亮的关键词
        keywordstyle=\bfseries % 设置关键词样式为加粗
    ]
DOMAIN_INTENT_LIST =
- Car Control: Car System Control, Car Body Control
- Map: Navigation, Provide Address, Check Traffic
[...]

SLOT_LIST = 
- Car Control-Object: Specific objects for car control, for example, air conditioning, interior lights, reading lights.
- Map-Map Size: The zoom level of the map, for example, minimum, zoom out.
[...]

SYSTEM_PROMPT_TEMPLATE =
You are an expert in SLU for automotive cabin systems.
Your task is to identify all the slots and their corresponding values from the user's query for slot filling.

You need to follow these rules:
1.  Identify Multiple Semantic Frames: A single user query may contain multiple independent intents. You need to generate a corresponding semantic structure for each intent.
2.  Strict Matching: The identified domain, intent, and slot names must be strictly selected from the list below. 

[...]

Available Domain-Intent List:
{DOMAIN_INTENT_LIST}

Available Slot List:
{SLOT_LIST}
    \end{lstlisting}
    \end{tcolorbox}
    \caption{In-context learning prompt template for jointly SLU task. The intent lists, slot lists, and format are partially omitted for brevity.}
    \label{fig:my_prompt_example}
\end{figure}

% - Map-Operation: Operations performed on the map, for example, navigation, display, search.

\begin{table*}[tbp]
    \centering
    \renewcommand{\arraystretch}{1.2} % Adjusts the row spacing for better readability
    % Use p{width} columns to allow text to wrap
    \begin{tabular}{p{2cm} p{7cm} p{7cm}}
        \toprule
        % -- Table Header --
        \textbf{Speech Query} & \textbf{Model Prediction} & \textbf{Ground Truth} \\
        \midrule
        
        % -- Row 1 --
        Close the car window & 
        \{Intent1: Car Control: [\{value: Body Control, name: intent\}, \{name: object, value: \textcolor{blue}{window}\}, \{name: action, value: close\}]\} & 
        \{Intent1: Car Control: [\{value: Body Control, name: intent\}, \{name: object, value: \textcolor{blue}{car window}\}, \{name: action, value: close\}]\} \\
        \midrule

        % -- Row 2 --
        Turn on the rear windshield defroster & 
        \{Intent1: Car Control: [\{value: Body Control, name: intent\}, \{name: action, value: turn on\}, \{name: position, value: rear\}, \{name: function, value: \textcolor{blue}{defrost}\}]\} & 
        \{Intent1: Car Control: [\{value: Body Control, name: intent\}, \{name: action, value: turn on\}, \{name: position, value: rear\}, \{name: function, value: \textcolor{blue}{windshield defrost}\}]\} \\
        \midrule

        % -- Row 3 --
        I want to listen to Jay Chou's representative works & 
        \{Intent1: Music: [\{name: action, value: \textcolor{blue}{I want to listen}\}, \{name: artist, value: Jay Chou\}, \{name: album, value: representative works\}, \{value: Play Music, name: intent\}]\} & 
        \{Intent1: Music: [\{name: action, value: \textcolor{blue}{listen}\}, \{name: artist, value: Jay Chou\}, \{name: object, value: representative works\}, \{value: Play Music, name: intent\}]\} \\
        \bottomrule
    \end{tabular}
    
    \caption{Comparison between fine-tuned Qwen2.5-Omni-7B predictions and ground truth. All data shown are English translations of the original Chinese. The differences are shown in \textcolor{blue}{blue}.}
    \label{tab:query_comparison}
    %\vspace{-5pt}
\end{table*}

\subsection{In-Context Learning Results}

\textbf{LLMs and LALMs are capable of performing SLU tasks through in-context learning.} Table \ref{tab:in_context_learning} shows the performance of LLMs and LALMs under direct inference or in-context learning. For the IC results, our findings are similar to \cite{he2023can}, namely that LLMs have the potential to complete the task using the in-context learning paradigm, and we extend this conclusion to the latest LALMs. For the SF results, we were surprised to find that with the carefully designed prompt shown in Figure \ref{fig:my_prompt_example}, LLMs and LALMs were also able to provide correct answers. For text and speech inputs, our system achieved F1 scores of up to 55.09\% and 47.38\%, respectively, which is significantly higher than 13.35\% reported in \cite{he2023can}. However, even with the best-performing models like Qwen3-32B or GPT-4o-Audio, their Overall Accuracy did not exceed 15\%, which reflects the challenges of the MAC-SLU dataset.

\textbf{E2E LALMs can achieve performance comparable to that of pipeline systems.} For speech input, Qwen2.5-Omni-7B slightly surpasses the pipeline system of a similar size (Whisper + Qwen3-8B), with advantages of 2\% in the IC task and 1\% in the SF task. This is attributed to the avoidance of ASR transcription error propagation in E2E SLU. However, when compared to the larger Qwen3-32B, the performance of Qwen2.5-Omni-7B still lags, indicating substantial potential for further performance improvements by scaling LALMs to larger sizes in the future. Furthermore, the closed-source GPT4o-Audio and Gemini-2.5-Flash demonstrate superior performance to open-source LALMs in the overall accuracy.

\subsection{SFT Results}

\textbf{LLMs and LALMs that undergo in-domain SFT exhibit performance superior to methods based on in-context learning.} Table \ref{tab:sft} presents the performance of LLMs and LALMs after being fine-tuned on the training set, where all models show significant improvements. SFT boosted Qwen2.5-Omni-7B's performance over in-context learning, with increases of 29\% in IC accuracy, 39\% in SF F1 score, and 47\% in overall accuracy. Currently, the most effective method for achieving optimal performance on SLU tasks with LLMs and LALMs remains fine-tuning on the training set. Enabling LLMs and LALMs agents to autonomously perform SLU tasks through in-context learning continues to pose a considerable challenge.

\begin{table}[t]
    \centering
    \renewcommand{\arraystretch}{1.2} % 设置表格行间距
    
    \begin{tabular}{l ccc}
        \toprule
        % -- 表头行 --
        \textbf{Model} & \textbf{Acc} & \textbf{F1} & \textbf{Overall Acc} \\
        \midrule
        
        % -- NLU 部分 --
        Qwen3-8B & 90.91 & 84.69 & 60.73 \\
        \midrule

        % -- ASR + NLU 部分 --
        Paraformer + Qwen3-8B & 88.92 & 79.09 & 47.18 \\
        Whisper + Qwen3-8B & 82.42 & 70.58 & 35.45 \\
        \midrule
        
        % -- E2E SLU 部分 --
        % 注意：根据您的要求，这里使用了 E2E SLU，尽管图片上写的是 SLU
        Qwen2-Audio-Instruct & 86.46 & 79.46 & 49.87 \\
        Qwen2.5-Omni-3B & 89.98 & 82.86 & \textbf{56.56} \\
        Qwen2.5-Omni-7B & \textbf{91.24} & \textbf{83.02} & 55.60 \\
        MiniCPM-o-2\_6 & 88.98 & 81.26 & 51.87 \\
        Phi-4-Multimodal-Instruct & 81.69 & 74.12 & 37.97 \\
        \bottomrule
    \end{tabular}
    
    \caption{SFT results for LLMs and LALMs on MAC-SLU dataset. The ASR model was not fine-tuned, while the Qwen3-8B was fine-tuned on the NLU task. The best results are shown in \textbf{bold}.}
    \label{tab:sft}
    \vspace{-5pt}
\end{table}

% \begin{table}[t]
%     \centering
%     \renewcommand{\arraystretch}{1.2} % 设置表格行间距
    
%     \begin{tabular}{l ccc}
%         \toprule
%         % -- 表头行 --
%         \textbf{Model} & \textbf{Acc} & \textbf{F1} & \textbf{Overall Acc} \\
%         \midrule
%         Qwen2.5-Omni-7B (ICL) & 54.36 & 45.56 & 9.73 \\
%         Qwen2.5-Omni-7B (SFT) & 91.24 & 83.02 & 55.60 \\
%         \bottomrule
%     \end{tabular}
    
%     \caption{Supervised Fine-tuning (SFT) and In-Context Learning (ICL) results for Qwen2.5-Omni-7B on MAC-SLU dataset.}
%     \label{tab:sft}
% \end{table}

\textbf{Error propagation in pipeline systems significantly degrades model performance.} When processing text queries (CER=0\%), the Qwen3-8B model achieves an overall accuracy of 60.73\%, the highest among all SFT models. However, when integrated into a pipeline system, the performance of Qwen3-8B deteriorates by over 13\% with transcripts from Paraformer (CER=3.64\%) and by more than 25\% with transcripts from Whisper (CER=10.40\%). Inaccurate transcriptions from ASR substantially reduce the efficacy of LLMs on SLU tasks, resulting in final performance that is markedly inferior to that of LALMs.

\subsection{Qualitative Analysis}

\textbf{A significant portion of observed errors arises because model outputs, while semantically correct, are phrased differently from the label.} Our case study in Table \ref{tab:query_comparison} shows that in most instances, the models' responses convey the intended meaning but are penalized for lexical variation. For the SLU task, which aims to extract user semantics for downstream applications, these outputs are functionally correct and would be accepted by human evaluators. Therefore, standard SLU metrics relying on exact string matching likely underestimate the true intent comprehension capabilities of LALMs.

\section{Conclusion}

This paper introduced MAC-SLU, a novel Chinese multi-intent SLU dataset for automotive cabin scenarios, addressing the lack of diversity and complexity in existing datasets. Building upon this dataset, we established the unified benchmark for open-source LLMs and LALMs on SLU tasks. Our experiments demonstrate that while in-context learning shows potential, in-domain SFT remains crucial for optimal performance. Furthermore, E2E LALMs achieve performance comparable to traditional pipeline methods with LLMs by effectively mitigating ASR error propagation. Future work should focus on enhancing the in-context learning capabilities of models and exploring more semantically-aligned evaluation metrics.

\bibliographystyle{IEEEbib}
\bibliography{main}

\end{document}